\documentclass[]{style}

\usepackage{graphicx}
\usepackage{svg}
\usepackage{float}
\usepackage{wrapfig}
\usepackage{subcaption}
\usepackage{caption}
\usepackage{adjustbox}
\usepackage{tikz}
\usetikzlibrary{positioning, fit}

\usepackage{array}
\usepackage{booktabs}
\usepackage{tabularx}
\usepackage{multirow}
\usepackage{makecell}
\usepackage{colortbl}

\usepackage[table]{xcolor}
\definecolor{headergray}{gray}{0.90}
\definecolor{rowgray}{gray}{0.95}
\definecolor{agentColor}{RGB}{0, 51, 153}
\definecolor{assistColor}{RGB}{128, 0, 128}
\definecolor{failRed}{RGB}{200, 0, 0}
\definecolor{jsonKey}{RGB}{0, 100, 0}
\definecolor{strColor}{RGB}{42, 0, 255}
\definecolor{codebg}{RGB}{245,246,247}
\definecolor{codeframe}{RGB}{200,200,200}

\usepackage{amsmath}
\usepackage{amssymb}
\usepackage{mathtools}
\usepackage{amsthm}

\usepackage{fontawesome5}
\usepackage{inconsolata}
\usepackage{microtype}

\usepackage{enumitem}
\usepackage{ragged2e}
\usepackage{comment}

\usepackage{listings}
\lstdefinestyle{chattemplate}{
  basicstyle=\ttfamily\small,
  breaklines=true,
  breakatwhitespace=true,
  showstringspaces=false,
  columns=fullflexible,
  frame=none,
  xleftmargin=0pt,
  aboveskip=0pt,
  belowskip=0pt
}

\usepackage[most]{tcolorbox}
\tcbuselibrary{listings,breakable,skins}

\usepackage[toc,page,header]{appendix}

\usepackage{hyperref}

\theoremstyle{plain}

\theoremstyle{definition}

\theoremstyle{remark}

\newcommand{\secref}[1]{Sec.~\ref{#1}}

\newcommand{\blue}[1]{\textcolor{blue}{#1}}

\newcommand{\figref}[1]{Fig.~\ref{#1}}

\newcommand{\figpanelref}[2]{Fig.~\ref{#1}#2}

\title{SCALE: Towards Whole-Slide Counterfactual Explanations\\in Cancer Histopathology}

\renewcommand{\authorlist}{%
  \authorfont\sffamily\bfseries
  Weiyi Qin$^{1,3}$,
  Yingci Liu-Swetz$^{2}$,
  Shiwei Tan$^{3}$,
  Yi Wang$^{1,3}$,
  Mohammad Alexanderani$^{4}$,
  Dimitrios Korentzelos$^{4}$,
  Peizi Li$^{5}$,
  Soo Yeon Sohn$^{5}$,
  Xintong Wang$^{5}$,
  Hao Peng$^{1}$,\\
  Yue Guo$^{1}$,
  Yuzhe Yang$^{6}$,
  Han Zhao$^{1}$,
  Hao Wang$^{1,3}$%
}

\affiliation[1]{University of Illinois Urbana-Champaign}
\affiliation[2]{University of Pennsylvania}
\affiliation[3]{Rutgers University}
\affiliation[4]{University of Pittsburgh}
\affiliation[5]{Mount Sinai Hospital}
\affiliation[6]{University of California, Los Angeles}

\abstract{
Accurate cancer diagnosis depends on histopathological evaluation, in which tumor morphology drives disease classification and therapeutic decision-making. While deep learning models have demonstrated strong performance in predicting diagnostic and prognostic properties from histopathology images, their clinical adoption remains limited by a lack of interpretability. Unlike conventional medical images where diagnostic findings are spatially localized, histopathological images are characterized by complex textural patterns distributed across the entire whole slide image (WSI), making it difficult to explain model predictions through region-based localization alone. Counterfactual explanations offer a promising alternative by asking ``what changes in the image could have altered the model prediction.'' To achieve this, we propose {SCALE}, a framework for generating such counterfactual explanations for histopathological DL models. Across nine counterfactual directions spanning multiple cancer types, SCALE generates counterfactual explanations that consistently shift model predictions toward the intended diagnostic target while maintaining biological feasibility. In a blinded review by pathologists, SCALE demonstrated superior morphological plausibility compared to baseline methods (win rate of 91.7\%) and robust counterfactual effectiveness (target consistency of 96.7\%). These results suggest that SCALE generates morphologically plausible and effective counterfactual explanations that can be applied to improve the interpretability of deep learning
models for histopathology.
}
\checkdata[\faDesktop~Website]{\url{https://scale-pathology.github.io/}}
\checkdata[\faEnvelope~Contact]{weiyiqin@illinois.edu, hwangml@illinois.edu}

\begin{document}

\maketitle

\section{Introduction}
\label{sec:intro}

Deep learning models in digital pathology can predict a wide range of diagnostic and molecular properties from whole-slide images (WSIs), including tumor type, histologic subtype and biomarker status \citep{coudray2018classification,campanella2019clinical,kather2019deep}. A widely used paradigm for WSI prediction is attention-based multiple-instance learning (MIL), which aggregates information from many tissue regions (e.g., tiles) and assigns greater weight to regions that contribute strongly to the slide-level prediction \citep{lu2021dataefficient}. These weights can be visualized as an attention map, highlighting regions emphasized by the model for its prediction. Yet these attention maps primarily indicate \emph{where} predictive evidence is concentrated, without revealing \emph{what} morphological features drive the decision~\citep{jamshidi2026beyond,qin2026clearhpv}. Because relevant morphology can be distributed across large tissue regions or even the whole slide, localization alone often fails to provide an effective explanation of model behavior. 

\begin{figure}[!t]
\centering
\includegraphics[
    width=1\textwidth,
]{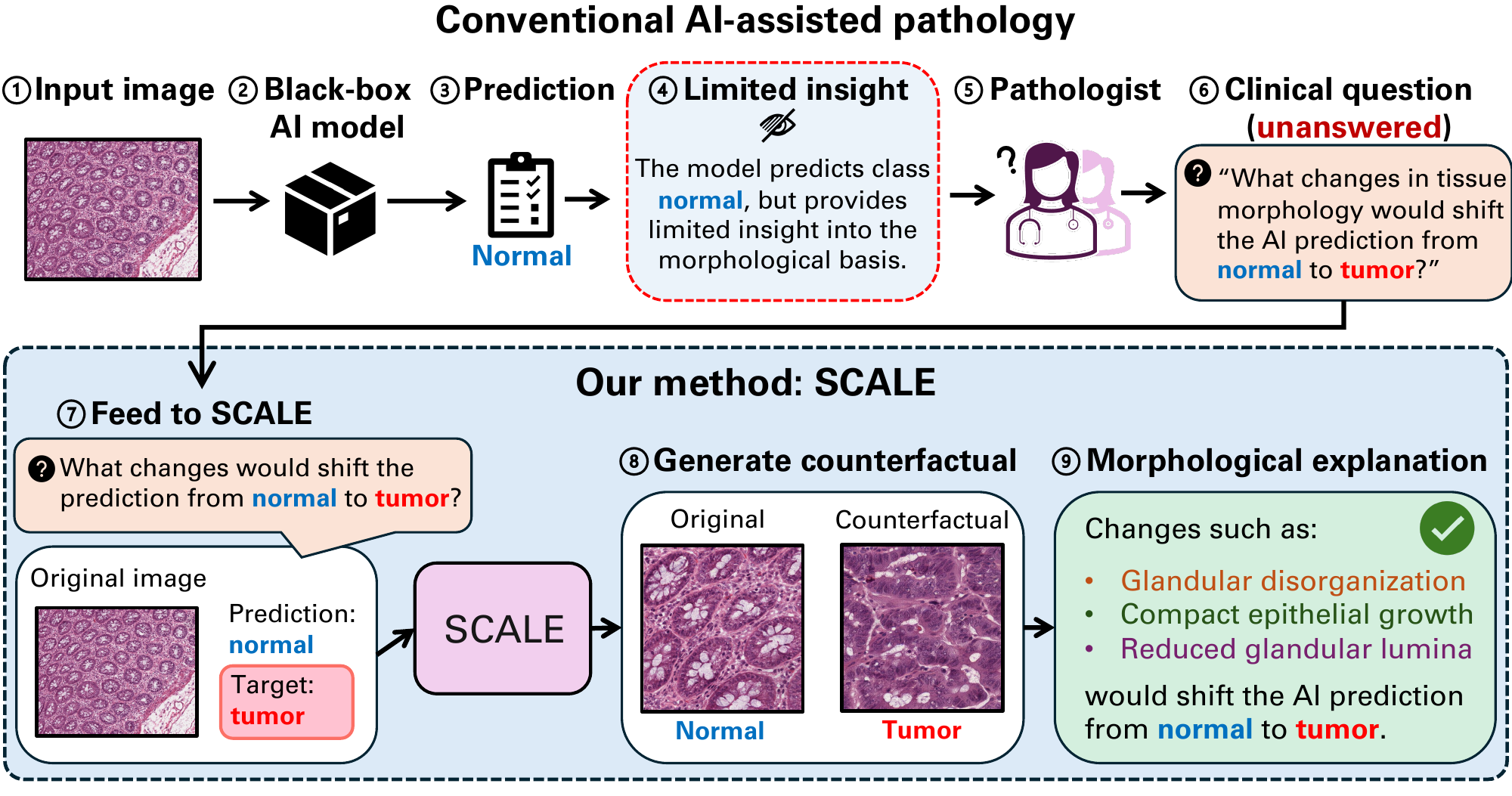}
\caption{
{\textbf{Potential integration of SCALE into AI-assisted pathology workflows.}
In conventional AI-assisted pathology, a black-box AI model provides a prediction from a histopathology image but offers limited insight into the morphological basis of its decision.
SCALE enables pathologists to investigate this decision by asking what changes in tissue morphology would shift the AI prediction from the original class to an alternative class (e.g., from ``normal'' to ``tumor'').
Given the original image, model prediction and specified counterfactual target, SCALE generates a counterfactual image.
Comparison of the original and counterfactual images reveals morphological changes associated with the prediction shift, i.e., ``glandular disorganization'', ``compact epithelial growth'', and ``reduced glandular lumina''. 
This workflow illustrates how SCALE could support pathologists in understanding and critically evaluating AI predictions through visually interpretable morphological interventions. Additional examples of class-associated morphological concepts across histologic subtypes, tumor grades and molecular statuses are shown in Fig.~\ref{fig:SAE}f.
}}
\label{fig:pipeline}
\end{figure}

\begin{figure}[!t]
\centering
\includegraphics[width=\textwidth]{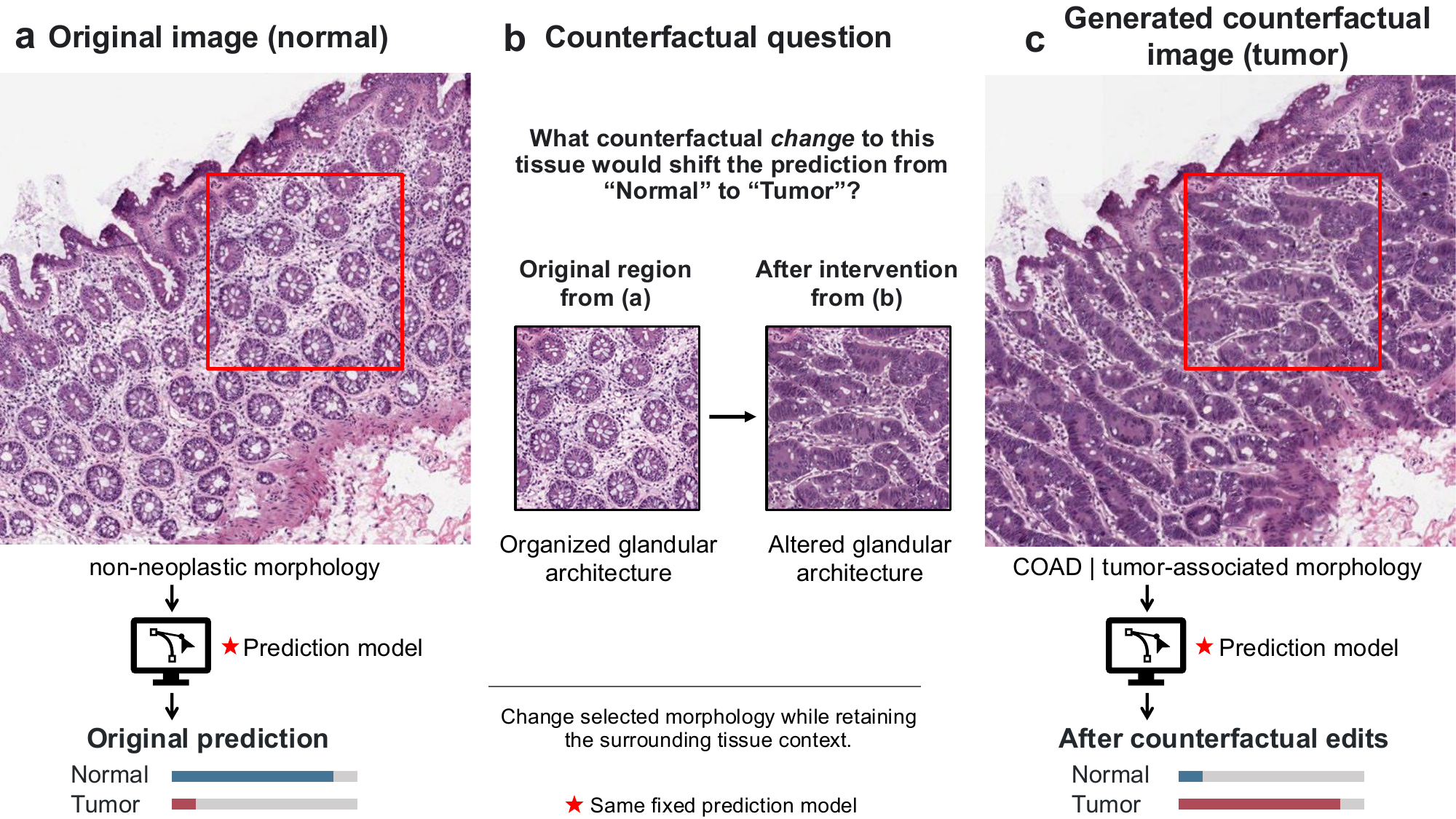}
\caption{
{\textbf{Conceptual illustration of model-directed counterfactual explanations in computational pathology.}
\textbf{a, Original image and model prediction.}
An original colorectal histopathology image is evaluated by a trained prediction model, which predicts the normal class. The red box highlights a region containing non-neoplastic glandular morphology.
\textbf{b, Counterfactual question.}
Counterfactual explanations investigate what hypothetical morphological changes to the original tissue would alter the model's prediction. The two images show the red-boxed regions from a and c side by side, illustrating the change from organized to altered glandular architecture.
\textbf{c, Counterfactual image and prediction shift.}
The generated counterfactual image is evaluated using the same fixed prediction model as in a, illustrating a prediction shift from normal toward tumor. The red box marks the spatially corresponding region in the counterfactual image. Comparing the model's responses before and after the morphological intervention illustrates how counterfactual explanations relate tissue changes to model predictions. 
}}
\label{fig:conceptual}
\end{figure}

{A more informative explanation would instead reveal how histologically plausible changes in morphology alter the model prediction.
As illustrated in \figref{fig:pipeline} {(top)}, rather than passively receiving an AI prediction with limited insight into its morphological basis, a pathologist could actively investigate the decision by asking what changes in tissue morphology would shift the model's prediction from one diagnostic class to another.
{As shown in \figref{fig:pipeline} (bottom), counterfactual explanations}\footnote{We use the term ``counterfactual'' in the model-explanation sense to denote a targeted input or representation intervention used to examine changes in model predictions, rather than the ``counterfactual'' defined by a formal causal model \cite{pearl2009causality}.}
provide a means to answer such questions by making prediction-altering morphological changes visually explicit. For example, if a model predicts a \emph{``negative'' diagnosis} for colon adenocarcinoma (COAD), conventional explanation methods that identify regions supporting this prediction may end up highlighting the entire WSI, rendering the explanation uninformative.
In contrast, counterfactual explanations reveal which changes in the WSI could have shifted the same model's prediction from \emph{``negative''} to \emph{``positive''} {(see \figref{fig:conceptual} for a more detailed illustration)}, allowing pathologists to examine the tissue features the model associates with alternative pathological states.
Showing whether predictions respond to pathologically plausible morphology can help assess whether model decisions rely on meaningful tissue features rather than spurious visual cues, providing a stronger basis for evaluating model trustworthiness \cite{zigutyte2026counterfactual}.
Realizing this potential at the whole-slide level, however, requires counterfactual methods that can not only identify prediction-relevant morphology but also generate plausible morphological changes at the large spatial scale over which that evidence is distributed.}

Existing interpretability and counterfactual methods satisfy only subsets of these requirements. Representation- and concept-based methods can recover interpretable morphological patterns associated with model predictions \citep{lu2021dataefficient,qin2026clearhpv}, but remain largely associational and do not show how the corresponding tissue morphology would need to change to produce a different outcome. Slide-level counterfactual approaches can manipulate tissue instances or representations \citep{kaczmarzyk2024hippo}, but are \emph{not generative} and therefore cannot synthesize the corresponding morphological transformations. Generative counterfactual methods can instead produce prediction-altering histopathology images \citep{zigutyte2026counterfactual,munoz2026histoxplaining}, but typically operate on localized image regions, e.g., on individual small tiles, and do not address how such interventions can be coherently propagated across larger tissue extents. Thus, generating counterfactuals that are simultaneously morphologically interpretable, spatially targeted, and scalable to the whole-slide level remains an open challenge.

Developing such a framework requires addressing three coupled technical problems: \emph{what} morphology to change, \emph{where} to apply the change, and \emph{how} to propagate it across the WSI.
First, a counterfactual intervention must specify \emph{what} morphology to change in an interpretable manner rather than perturbing arbitrary features in a high-dimensional representation. Second, because WSIs contain extensive and heterogeneous tissue, the framework must determine \emph{where} an intervention should be applied so that the edited regions are relevant to the model prediction. Third, the gigapixel scale of WSIs makes whole-slide generation impractical in a single pass, requiring a strategy for determining \emph{how} localized morphological changes can be coherently propagated across larger tissue regions while preserving continuity between neighboring edits.

\begin{figure}[!htbp]
\centering
\includegraphics[width=\textwidth]{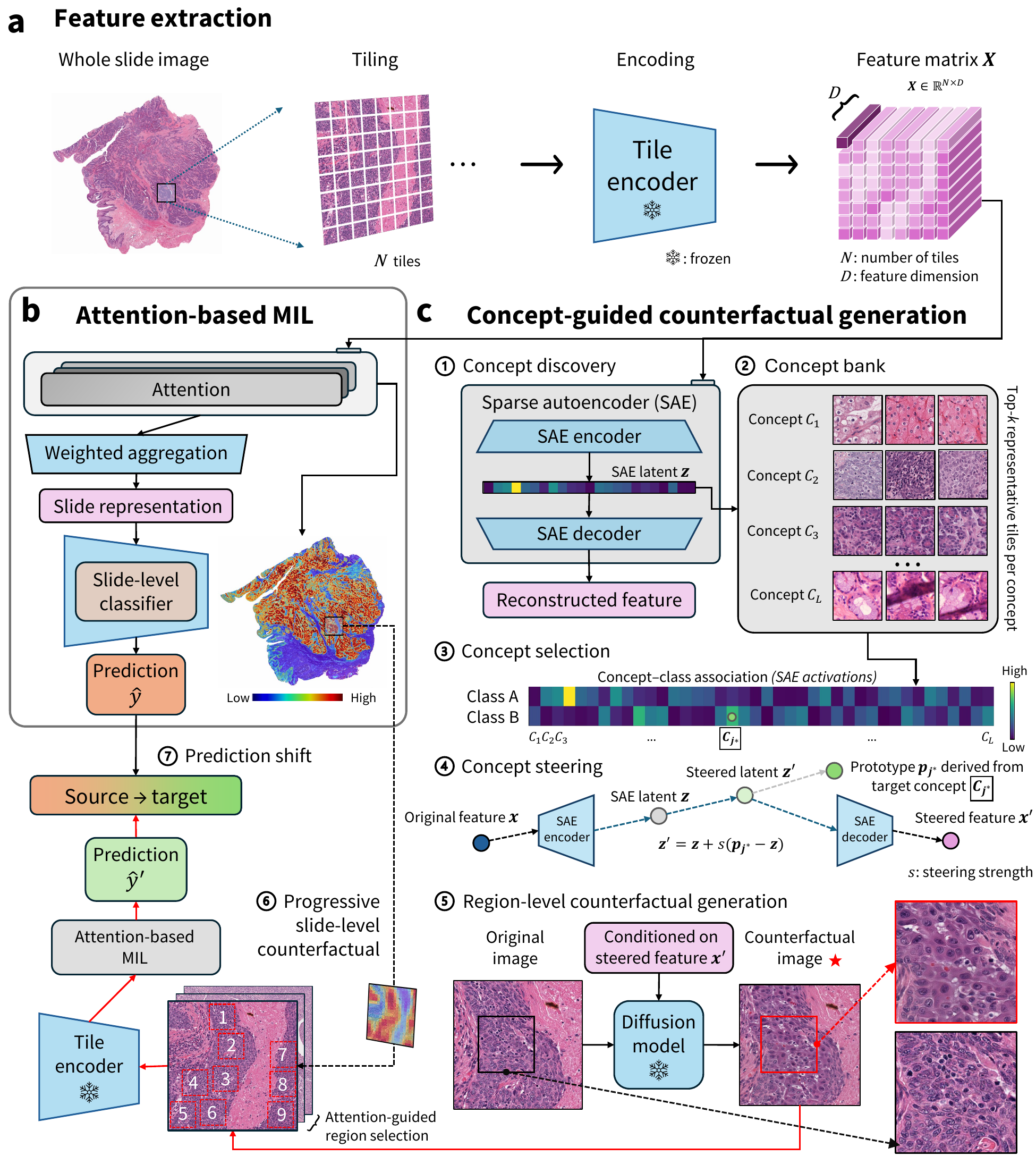}
\caption{
\textbf{Overview of SCALE.}
\textbf{a, Feature extraction.} Whole-slide images are partitioned into tiles and encoded by a pretrained tile encoder to obtain {tile}-level feature embeddings. Each slide is represented as a feature matrix \(X \in \mathbb{R}^{N \times D}\), where \(N\) denotes the number of tiles and \(D\) denotes the feature dimension. 
\textbf{b, Attention-based multiple instance learning.} Tile-level feature embeddings are aggregated using an attention-based MIL model to produce slide-level predictions. The model assigns attention weights to individual tiles, highlighting regions that contribute most to the prediction and enabling attention-guided region selection for generating counterfactual explanations. \textbf{c, Concept-guided counterfactual generation.} {An} SAE is trained on large-scale TCGA data to discover a reusable concept bank from pathology feature representations, with each concept grounded by prototype examples. Concept-class association identifies task-relevant concepts, and a target concept \(c_{j*}\) is selected for intervention. The original feature \(x\) is steered in the SAE latent space to obtain a modified feature \(x'\); a diffusion model is then conditioned on \(x'\) to generate region-level (e.g., tile-level) counterfactual explanations. Progressive editing integrates these local edits across the slide to produce a coherent slide-level counterfactual with the desired prediction shift. 
}
\label{fig:framework}
\end{figure}

To address these challenges, we introduce Scalable Counterfactuals through Attention-guided Latent Editing in Pathology (SCALE), a pathology-grounded framework that unifies \emph{what} morphology to change, \emph{where} to intervene, and \emph{how} to propagate counterfactual edits across tissue, enabling interpretable counterfactual generation that scales from local regions to the whole-slide level (details in \figref{fig:framework} and~\secref{sec:overview}). Across nine counterfactual directions spanning six cancer types and multiple biological endpoints, including normal-to-tumor transitions, human papillomavirus (HPV)-associated morphology, and prostate cancer growth patterns, SCALE produces coherent, concept-aligned morphological transformations {localized tissue regions to whole-slide images}, while consistently moving model predictions toward the intended targets. {Blinded assessment by six pathologists further evaluates the relative morphological plausibility and target consistency of the generated counterfactuals, complementing classifier-based evaluation with expert assessment of tissue morphology.} Together, these results establish SCALE as a unified framework for interpreting whole-slide pathology models through explicit morphological counterfactuals.

\section{Results}

\subsection{Overview of SCALE}\label{sec:overview}

\figref{fig:framework} summarizes SCALE as a closed-loop counterfactual explanation framework for identifying and generating morphological changes that alter whole-slide model predictions. It consists of three main stages. 

Each WSI is first partitioned into tiles, which are encoded using a pretrained tile encoder to obtain tile-level feature embeddings, as shown in \figref{fig:framework}a. A slide is represented by a feature matrix \(X \in \mathbb{R}^{N \times D}\), where \(N\) is the number of tiles and \(D\) is the feature dimension. 

These tile-level embeddings are then aggregated using an attention-based MIL model to produce slide-level predictions, as shown in \figref{fig:framework}b. The learned attention weights quantify the contribution of individual tiles to the prediction and are used to identify salient regions for subsequent counterfactual explanation. 

Most importantly, as shown in \figref{fig:framework}c, we introduce a concept-guided counterfactual generation framework built on an SAE trained on large-scale TCGA pathology data to learn a reusable bank of concepts from pathology feature representations, with each concept grounded using prototype examples. Concept-class associations are then used to identify task-relevant concepts and select a target concept \(c_{j^*}\) for intervention. Given an original tile feature \(x\), we steer its representation in the SAE latent space toward the target concept to obtain a modified feature \(x'\). A diffusion model conditioned on \(x'\) subsequently generates a tile-level counterfactual image corresponding to the desired concept intervention. Finally, progressive editing integrates these local (tile-level) counterfactual edits across salient regions of the slide, yielding a coherent slide-level counterfactual that induces the desired shift in the model prediction.

The generated counterfactual image (consisting of multiple tiles) is returned to the same downstream model to measure whether the morphological intervention shifts the prediction toward the intended target, closing the counterfactual loop (\figpanelref{fig:framework}{c}). 

Below, we first establish that SCALE's learned concepts provide interpretable morphological patterns for intervention, test whether SCALE's interventions alter model predictions and where they are most effective, and then evaluate the pathological validity and spatial scalability of SCALE's generated counterfactuals. Further implementation details for each component of the workflow are provided in the \secref{meth} section.

\subsection{Pan-cancer foundation-model representations yield interpretable, reusable morphological concepts}

\begin{figure}[!htbp]
\centering
\includegraphics[
    width=1\textwidth,
]{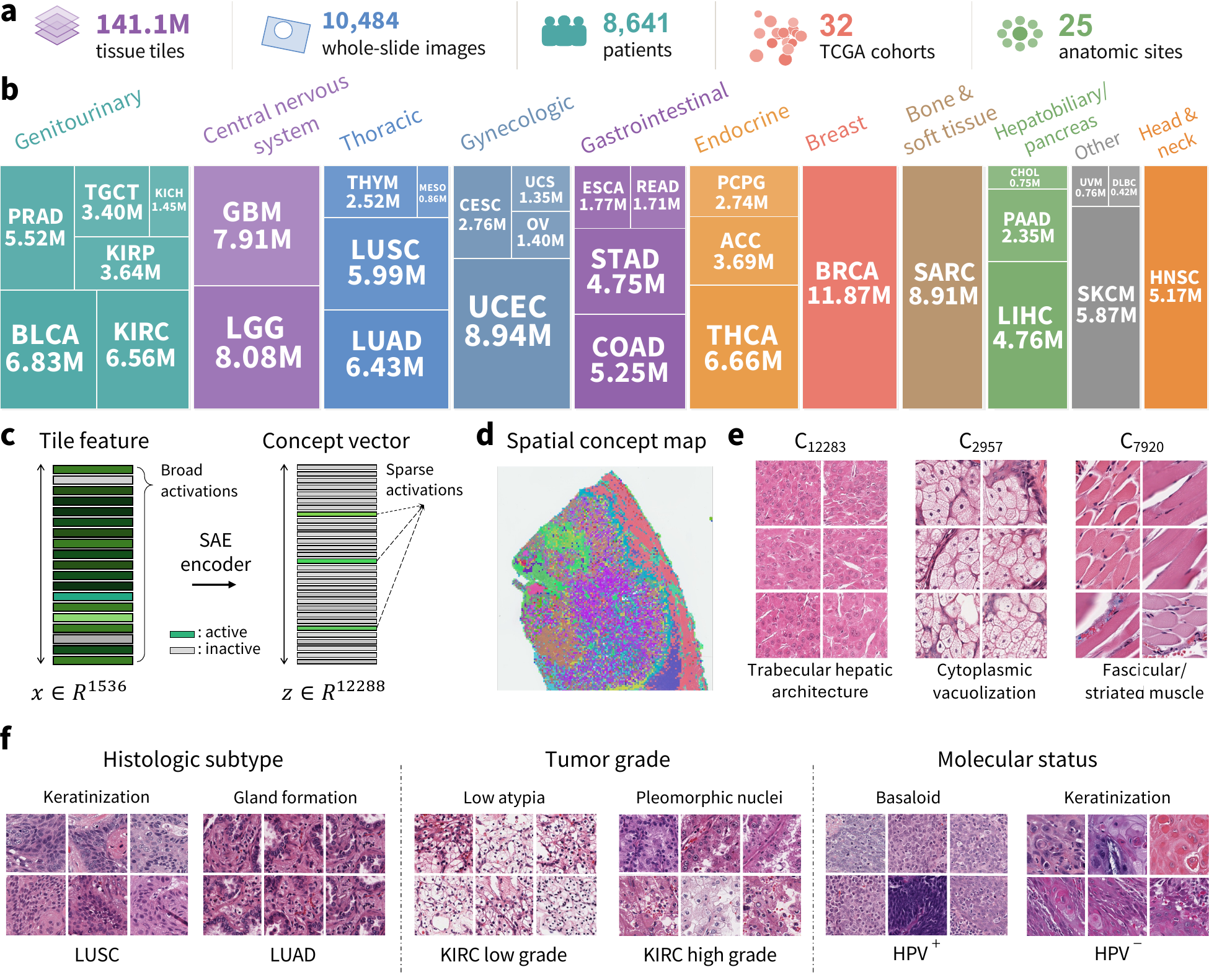}
\caption{
\textbf{Pan-cancer concept discovery and class-associated morphology.}
\textbf{a,} Sparse autoencoder (SAE) training data comprise 141.1 million tissue tiles from 10,484 whole-slide images and 8,641 patients across 32 The Cancer Genome Atlas (TCGA) cohorts and 25 anatomical sites. Tiles measure $256\times256$ pixels at $20\times$ magnification or 0.5 µm per pixel.
\textbf{b,} Tile contributions by cohort, grouped by anatomical system; labels report counts in millions.
\textbf{c,} The SAE maps a dense tile feature $x\in\mathbb{R}^{1536}$ to an overcomplete sparse concept vector $z\in\mathbb{R}^{12288}$. Colored and gray entries schematically indicate nonzero and zero activations, respectively.
\textbf{d,} Spatial concept map of an example tissue region. Each tile is colored by its highest-activation SAE concept; different colors denote distinct concepts.
\textbf{e,} Representative top-activating tiles for concepts $C_{12283}$, $C_{2957}$ and $C_{7920}$, illustrating trabecular hepatic architecture, cytoplasmic vacuolization and fascicular or striated muscle, respectively. Subscripts denote SAE concept indices.
\textbf{f,} Class-associated concepts across histologic subtype, tumor grade and molecular status: keratinization in LUSC and gland formation in LUAD; low atypia and pleomorphic nuclei in low- and high-grade KIRC; and basaloid morphology and keratinization associated with HPV$^{+}$ and HPV$^{-}$ head-and-neck squamous cell carcinoma, respectively. Galleries show representative top-activating tiles for the concept with the highest mean attention-weighted activation within each class. 
}
\label{fig:SAE}
\end{figure}

Applying SAE decomposition to pan-cancer pathology foundation-model representations yielded a reusable set of visually interpretable morphological concepts. To capture broad morphological diversity, we trained the SAE on foundation-model representations \citep{chen2024uni} from 141.1 million tissue tiles across 10,484 WSIs and 8,641 patients in 32 TCGA cohorts and 25 anatomical sites \citep{hoadley2018cellorigin} (\figref{fig:SAE}a). Tile contributions were broadly distributed across organ systems and cancer types, providing a diverse representation space from which shared morphological concepts could be identified (\figref{fig:SAE}b provides more detailed statistics).

The resulting sparse concepts captured recurring and spatially coherent tissue morphologies. Top-activating tiles for individual concepts showed recognizable tissue structures and cellular patterns, 
even though concept discovery does \emph{not} rely on predefined morphological categories or tile-level concept annotations (\figref{fig:SAE}c,e). 
Mapping dominant concept activations across WSIs further revealed spatially organized morphological domains within tissue (\figref{fig:SAE}d), indicating that the sparse representation recovered coherent morphological patterns from the underlying foundation-model feature space.

Importantly, the same concept space could be reused across distinct downstream prediction tasks and cancer types. Attention-weighted concept distributions differed between classes (\figref{fig:SAE}d), and strongly class-associated concepts exhibited distinct prototype morphologies across histologic subtype, tumor grade and viral-status settings (\figref{fig:SAE}f). For example, \figref{fig:SAE}f (left) shows that two histologic subtypes of lung cancer, lung squamous cell carcinoma (LUSC) and lung adenocarcinoma (LUAD), show different squamous and gland-forming morphologies, respectively, in their representative tiles. 
Similar examples include low nuclear atypia versus prominent nucleoli and pleomorphic nuclei in clear-cell renal cell carcinoma, shown in \figref{fig:SAE}f (middle), and basaloid versus keratinizing patterns associated with HPV status in head-and-neck squamous cell carcinoma, shown in \figref{fig:SAE}f (right). 
Together, these results show that pan-cancer foundation-model representations can be decomposed into visually grounded morphological concepts that are reusable across prediction tasks and provide interpretable targets for subsequent counterfactual intervention.

\subsection{Concept-guided counterfactual explanations reveal how morphology alters model predictions}

Concept-guided counterfactual explanations revealed morphological changes that systematically altered downstream model predictions across diverse pathology tasks. {Unlike attention-based interpretability, which only localizes prediction-relevant regions,} {SCALE performs counterfactual interventions by steering the original WSI (e.g., with HPV$^{-}$) toward morphology more associated with target label (e.g., HPV$^{+}$) and measures how the same prediction model responds to the resulting counterfactuals.} We evaluated nine counterfactual directions spanning breast, colorectal, kidney, lung, head-and-neck and prostate pathology, including normal-to-tumor transitions, bidirectional HPV-status transitions (between HPV$^{+}$ and HPV$^{-}$), and changes between prostate growth-pattern classes. Systematic evaluation was performed primarily on \(2048\times2048\)-pixel tissue regions to provide a common large-region setting across tasks. For these experiments, predictions were computed from each region's tile bag using the same task-specific MIL model, with all tiles affected by generation or blending re-encoded before evaluation.

\begin{figure}[!t]
\centering
\includegraphics[width=\textwidth]{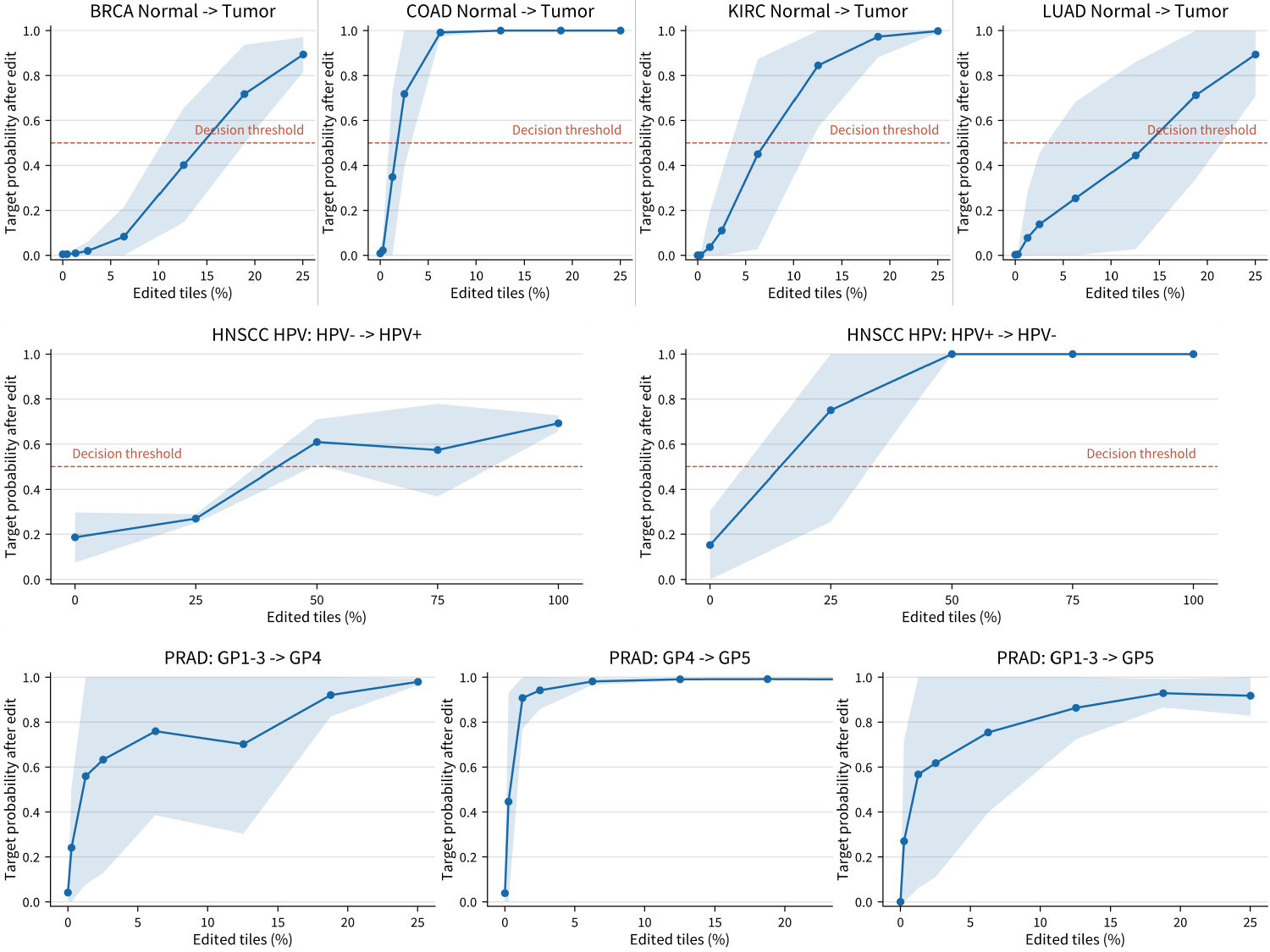}
\caption{
{\textbf{Whole-slide prediction responses to increasing counterfactual editing extent using SCALE across nine pathology tasks.}
Target-class probability as a function of the percentage of tissue tiles counterfactually edited across whole-slide images (WSIs). The evaluation included $n=134$ distinct WSIs from 101 TCGA cases across nine counterfactual task directions. For PRAD, Gleason patterns 1--3 (GP1--3) were grouped into a single class characterized by discrete, well-formed glands based on the historical Gleason morphological classification. Panels show, from left to right and top to bottom: BRCA normal-to-tumor, COAD normal-to-tumor, KIRC normal-to-tumor, HNSCC HPV$^{-}$-to-HPV$^{+}$, HNSCC HPV$^{+}$-to-HPV$^{-}$, LUAD normal-to-tumor, PRAD GP1--3-to-GP4, PRAD GP4-to-GP5 and PRAD GP1--3-to-GP5 editing. Horizontal red dashed lines indicate a target-class probability of 0.5 as a common reference level.}
}
\label{fig:benchmark}
\end{figure}

{Across nine counterfactual directions, target-class probability generally increased with the percentage of tissue tiles edited across WSIs, evaluated on 134 distinct WSIs from 101 TCGA cases (\figref{fig:benchmark}). However, the editing extent required to induce substantial prediction shifts varied considerably across tasks. Among the normal-to-tumor transitions, COAD and KIRC showed rapid responses, with target-class probabilities approaching 1.0 after editing approximately 6\% and 25\% of tissue tiles, respectively. In contrast, BRCA and LUAD exhibited more gradual increases, reaching approximately 0.9 at 25\% editing. HPV-status transitions also showed asymmetric responses: HNSCC HPV$^{+}$-to-HPV$^{-}$ reached a target-class probability near 1.0 at 50\% editing, whereas HPV$^{-}$-to-HPV$^{+}$ crossed the 0.5 decision threshold only after editing 50\% of tissue tiles and reached approximately 0.7 at 100\% editing. For PRAD, GP4-to-GP5 editing produced a particularly rapid response, exceeding 0.9 target-class probability after editing approximately 1\% of tissue tiles, while GP1--3-to-GP4 and GP1--3-to-GP5 transitions showed more gradual increases. 
Consistent target-directed prediction shifts were also observed in the complementary large-region evaluation using fixed editing budgets of 1--48 tiles , further demonstrating the effectiveness of SCALE across different spatial scales.
Together, these findings demonstrate that increasing the spatial extent of counterfactual interventions induces target-directed prediction shifts across both whole-slide and large-region settings, although the editing extent required varies substantially across pathology tasks.
}

\subsection{Spatial selection, steering strength and intervention extent provide complementary control}

\begin{figure}[!htbp]
\centering
\includegraphics[width=1\linewidth]{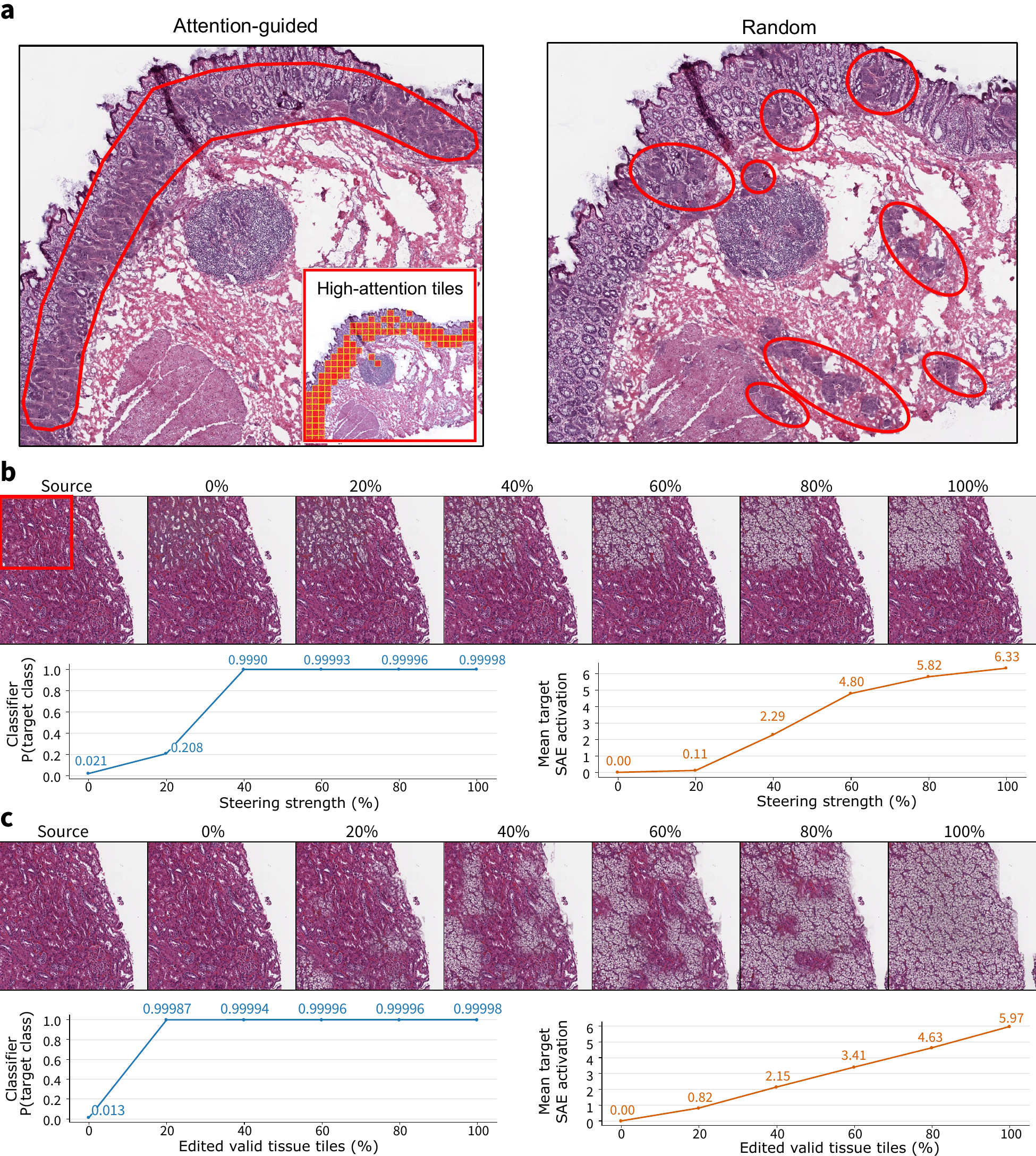}
\caption{
\textbf{Spatial selection, steering strength and intervention extent.}
\textbf{a,} Colorectal counterfactuals generated using attention-guided selection (left) or random selection (right) at a matched selected-tile budget. The inset highlights high-attention tiles selected for intervention. Red outlines indicate representative intervention areas, rather than exact affected-pixel masks.
\textbf{b,} Steering-strength sweep in a representative kidney region, with the selected area fixed (red box). Images show the source and counterfactuals at increasing steering strengths. Plots show region-level target-class probability (left) and mean target-concept SAE activation across the indicated 16 tiles (right).
\textbf{c,} Intervention-extent sweep in the same kidney region, with steering strength fixed and increasing percentages of valid tissue tiles selected for intervention. Plots show region-level target-class probability (left) and mean target-concept SAE activation across all 55 valid tissue tiles (right). 
}
\label{fig:attention-edit}
\end{figure}

Having established target-directed prediction responses, we next examined control over intervention location, steering strength\blue{,} and spatial extent. 
SCALE uses attention from the same MIL model whose prediction is being explained to prioritize tissue tiles for intervention. In the representative colorectal example (\figref{fig:attention-edit}a), attention-guided selection concentrated edits within high-attention, epithelium-rich tissue containing glandular structures directly relevant to the intended morphological transformation. Random selection at a matched selected-tile budget instead dispersed interventions across heterogeneous tissue compartments, including areas with relatively sparse epithelial content. This verifies that SCALE's attention-guided selection aligns the intervention with both the model's spatial weighting and the tissue structures where the requested morphological change could be meaningfully expressed. {Consistent with this qualitative example, attention-guided selection produced larger or equal target-probability shifts than random selection at matched selected-tile budgets across all nine intervention directions, with the largest advantage in directions where target-associated tissue was spatially sparse, such as the HPV$^{-}$-to-HPV$^{+}$ transition.}

To distinguish the effects of local steering (intervention) strength (for each tile) and spatial extent (i.e., how many tiles are edited), we varied each aspect independently in a representative region. 
For the steering-strength experiment, the steering (intervention) was restricted to the fixed region indicated by the red box while only steering strength was increased. Increasing steering strength from 0\% to 40\% increased target-class probability from 0.021 to 0.999 and mean target-concept SAE activation from 0 to 2.29; further increasing steering strength to 100\% raised target-concept activation to 6.33 while the classifier response remained near saturation (\figref{fig:attention-edit}b). Increasing the fraction of edited valid tissue (i.e., number of tiles) produced a similarly strong target-directed response: target-class probability increased from 0.013 without steering to 0.99987 at 20\% editing, while mean target-concept activation increased progressively from 0 to 5.97 as the edited fraction expanded to 100\% (\figref{fig:attention-edit}c). Together, these experiments separate two complementary dimensions of counterfactual control: how strongly target-associated morphology is introduced at each selected location and how broadly the intervention is distributed across tissue. SCALE therefore reveals not only what morphology can change a prediction, but also how strongly and how extensively that morphology must be represented for the prediction model to respond.

We also tested whether the selected SAE concept could be steered directly through a single latent dimension. Single-neuron steering often produced off-target changes or visible artifacts, while SCALE's prototype-guided steering yielded more coherent target-associated morphology. 
This supports using the full latent prototype rather than an isolated SAE dimension for counterfactual generation.

\subsection{SCALE generates target-associated counterfactual explanations from local regions to whole slides} \label{sec:counter-whole}

\begin{figure}[!htbp]
\centering
\includegraphics[width=0.9\textwidth]{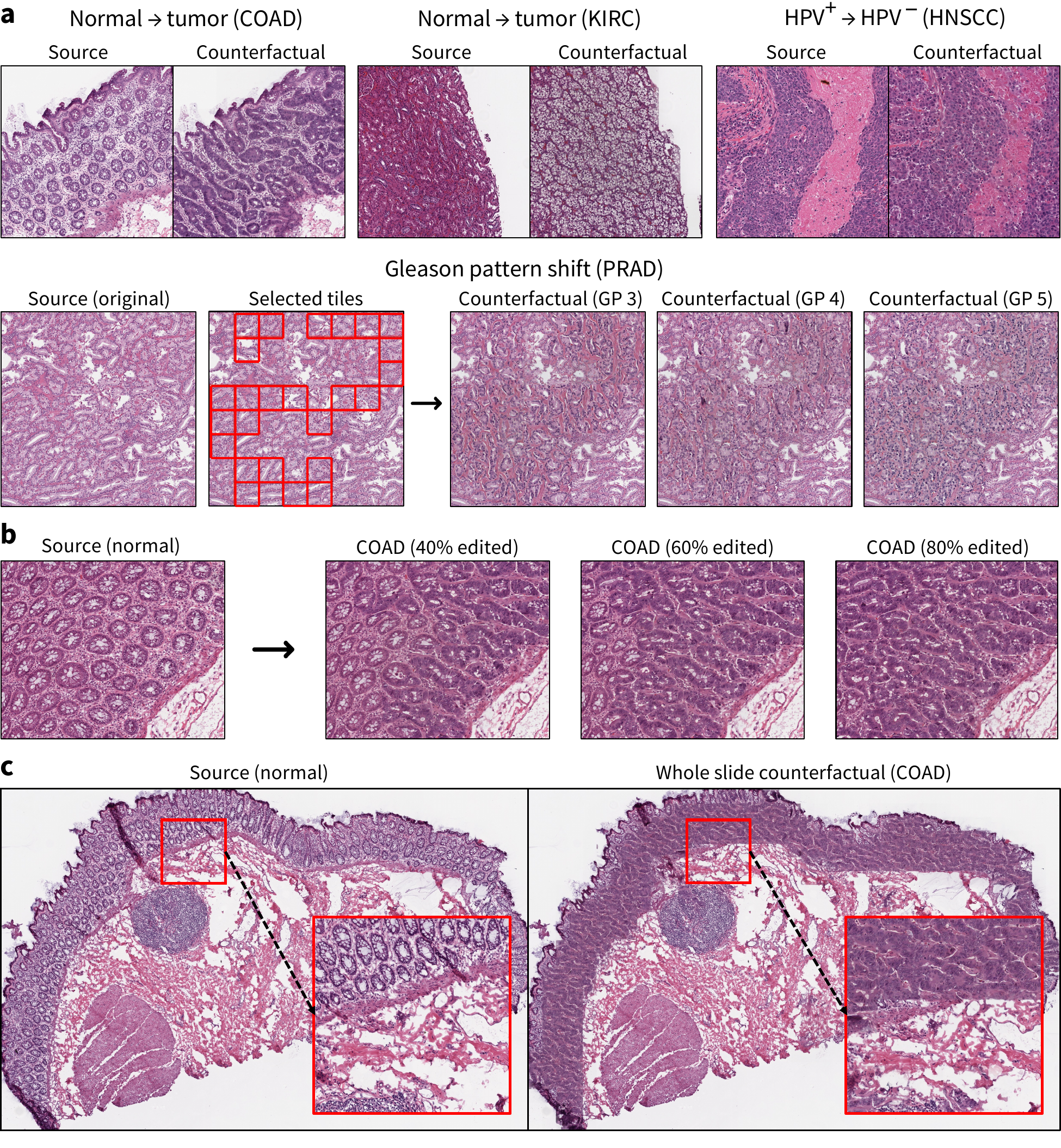}
\caption{
\textbf{Representative counterfactual explanations across pathology tasks from SCALE.}
\textbf{a,} Source and concept-guided counterfactual regions for colorectal (COAD) normal-to-tumor, kidney (KIRC) normal-to-tumor and HNSCC HPV$^{+}$-to-HPV$^{-}$ editing. For PRAD, the same source region is steered toward progressively higher-grade Gleason pattern (GP) targets, illustrating controlled changes in glandular architecture.
\textbf{b,} Colorectal normal-to-tumor counterfactuals generated while increasing the fraction of edited valid-tissue tiles from 40\% to 80\%. Increasing the edited fraction expands tumor-associated morphology across the tissue region, providing a spatial analogue of increasing local tumor involvement. Percentages denote the fraction of valid-tissue tiles selected for intervention.
\textbf{c,} Representative COAD normal-to-tumor whole-slide counterfactual. Progressive editing propagates the target-associated morphological intervention across spatially distributed tissue regions while retaining the overall slide context. Red boxes indicate the enlarged region.
Across all panels, counterfactual labels indicate intended intervention targets rather than definitive pathological classifications of the generated tissue.
}
\label{fig:qualitative}
\end{figure}

For a counterfactual to provide a meaningful morphological explanation of model behavior, the prediction shift should correspond to a recognizable and pathologically plausible alternative tissue state rather than an arbitrary image perturbation. As illustrated in \figref{fig:qualitative}a, SCALE produced target-directed prediction changes together with visible morphological transformations consistent with the corresponding pathological state. 
In {COAD}'s normal-to-tumor counterfactuals (\figref{fig:qualitative}a(left)), the regular organization of colonic crypts was replaced by more crowded and irregular gland-forming epithelium, consistent with the glandular architectural abnormalities characteristic of colorectal adenocarcinoma \citep{fleming2012colorectal,awan2017glandular}. In kidney tissue (\figref{fig:qualitative}a(middle)), editing toward the tumor class introduced nests of cells with clearer cytoplasm and altered epithelial architecture characteristic of clear-cell renal cell carcinoma \citep{nezami2025ccrcc}. HPV-positive head-and-neck tissue edited toward the HPV-negative class showed a shift from predominantly basaloid morphology toward more eosinophilic, squamous-differentiated and keratinizing features, consistent with established HPV-associated morphological differences in HNSCC \citep{Gillison2000JNCI,Ang2010NEJM,Linton2013BasaloidSCC} (\figref{fig:qualitative}a(right)). 
Importantly, these transformations remained organized at the cellular and tissue-architectural levels rather than appearing as arbitrary texture changes, supporting their interpretation as histologically plausible counterfactual alternatives and allowing them to serve as counterfactual explanations for the predictions (e.g., HPV-positive head-and-neck tissue). Pathological review indicated that these transformations remained histologically plausible and recognizable as the intended counterfactual alternatives.

Besides binary cases, SCALE also captured more graded morphological alternatives. In prostate tissue (\figref{fig:qualitative}a(bottom)), steering the same source region toward progressively higher-grade growth patterns produced corresponding changes in glandular architecture consistent with established Gleason pattern (GP) distinctions \citep{epstein2016contemporary,epstein2016isup} and the requested (higher-grade) target patterns. 
In colorectal tissue, increasing the fraction of edited valid-tissue tiles from 40\% to 80\% progressively expanded tumor-associated morphology across the region (\figref{fig:qualitative}b), demonstrating control over the spatial extent of the morphological explanation. Together, these results show that SCALE explains changes in model predictions through explicit, histologically plausible morphological alternatives, grounding the counterfactual explanation in tissue changes that are recognizable in pathological terms.

To assess the morphological quality and effectiveness of the generated counterfactuals beyond classifier outputs, six board-certified pathologists evaluated 30 cases spanning HPV-status, PRAD growth-pattern, and COAD tasks in a blinded assessment, yielding 60 reader-case assessments for each of two complementary questions. First, we assessed \textbf{morphological plausibility} by asking reviewers to compare SCALE with the evaluated baselines and rate which generated image appeared most histologically plausible. Second, we assessed \textbf{counterfactual target consistency} by asking whether the SCALE-edited image or its corresponding source image was more consistent with the morphology associated with the requested target class (for example, basaloid morphology for an HPV$^{+}$ target versus keratinizing morphology for an HPV$^{-}$ target, or architectural changes consistent with a transition from lower-grade prostate cancer growth patterns (GP1-3) toward GP4 or GP5).

{For \emph{morphological plausibility}, SCALE achieved a strict win rate of 91.7\% (55 of 60 reader-case assessments), receiving ratings higher than both single-neuron SAE steering and independent-window editing. Neither baseline achieved a strict win. 
For \emph{counterfactual target consistency}, 
pathologists selected SCALE counterfactual explanations over their corresponding source (original) images as more consistent with the requested target morphology in 58 of 60 reader-case assessments (96.7\%). These assessments covered 30 unique cases, each evaluated by two pathologists. Together, these results establish that SCALE produces counterfactual explanations that expert pathologists both accept as histologically realistic and recognize as expressing the requested target morphology, the two properties a counterfactual explanation must have to be trusted as evidence about model behavior.}

A central goal of SCALE is to extend localized counterfactual interventions from selected tissue regions to the spatial scale of whole-slide pathology. We therefore evaluated whether the same concept-guided, attention-directed editing strategy could produce coherent target-associated changes across a WSI while preserving the broader tissue context. 
In a representative COAD normal-to-tumor example, SCALE progressively targeted high-attention regions prioritized by the model and introduced tumor-associated glandular morphology without globally replacing or restructuring the slide (\figref{fig:qualitative}c). Enlarged views showed localized transitions from organized colonic glands toward more irregular tumor-associated glandular architecture, while surrounding tissue organization remained largely preserved. 

Importantly, these spatially distributed edits produced a corresponding shift in the whole-slide prediction. In the whole-slide analysis, {the classifier's predicted probability of the tumor class (target), averaged across all evaluated slides, }increased from approximately 0.02 to 0.72 after only about 50 edited tiles, crossing the 0.5 decision threshold, and reached approximately 0.99 after about 140 tiles. This shows that SCALE can accumulate localized, morphologically targeted interventions across an entire WSI to induce a controlled change in the slide-level prediction. This whole-slide setting demonstrates that the SCALE framework is not restricted to isolated image patches or fixed regions, but can scale counterfactual explanations to the spatial context in which computational pathology models operate.

\subsection{Progressive editing enables scalable counterfactual generation with reduced boundary artifacts}

\begin{figure}[!htbp]
\centering
\includegraphics[width=1\textwidth]{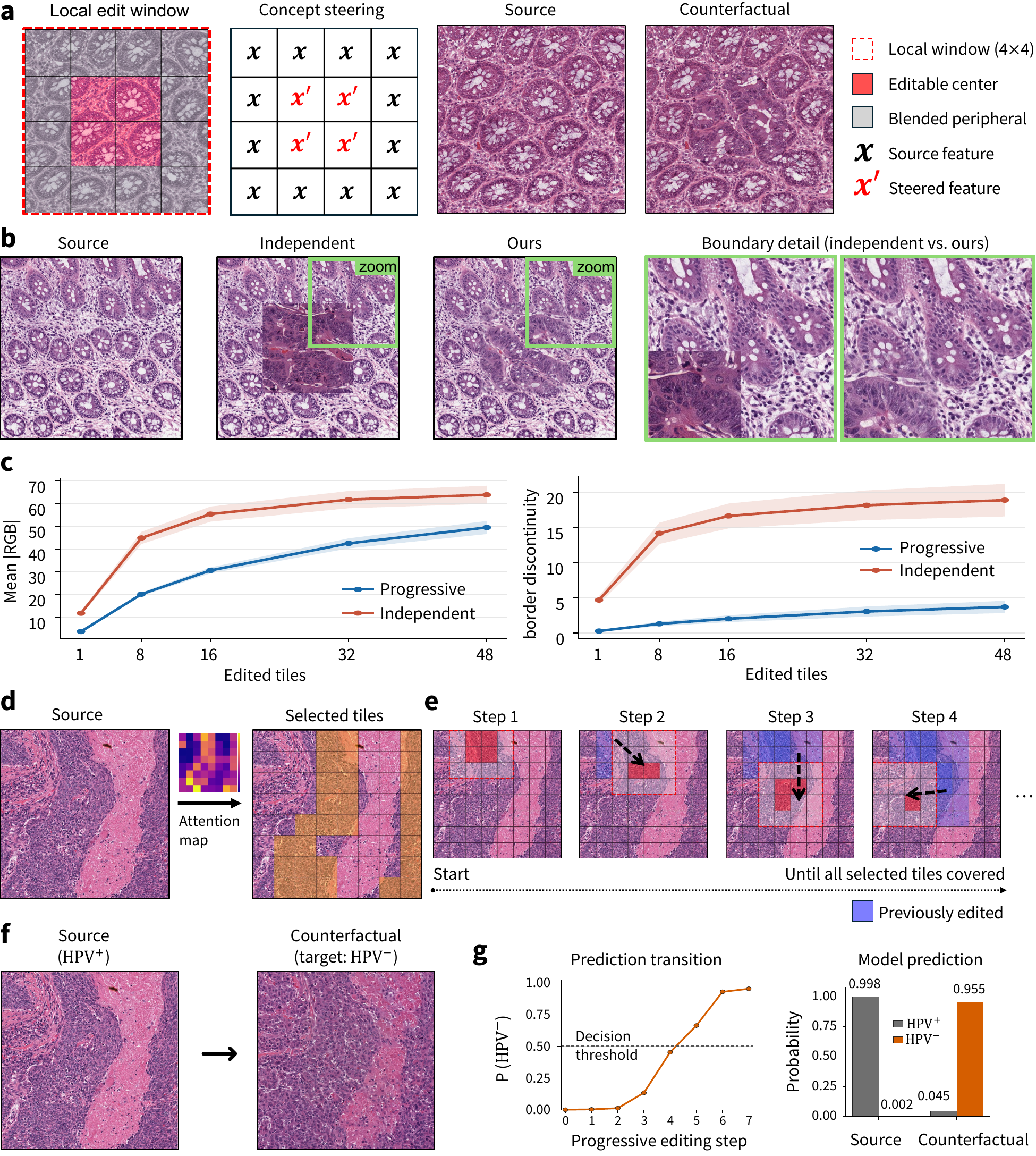}
\caption{
\textbf{Spatial continuity and prediction changes during progressive editing.}
\textbf{a,} Local editing uses overlapping $4\times4$-tile windows with a fully edited $2\times2$ center and a blended peripheral region (10\% generated, 90\% existing content).
\textbf{b,} Source tissue and counterfactuals generated by independent stitching or progressive editing (ours). Green boxes mark zoomed-in regions.
\textbf{c,} Mean absolute RGB change (\textbf{left}) and excess border discontinuity (\textbf{right}) across increasing selected-tile budgets (lower is better for both metrics). Lines indicate means, and shaded bands indicate the standard error of the mean across nine intervention directions ($n=9$).
\textbf{d--g,} Representative HPV$^{+}$-to-HPV$^{-}$ intervention.
\textbf{d,} Source region, attention map and selected tiles.
\textbf{e,} Progressive editing through overlapping windows.
\textbf{f,} Source and resulting counterfactual.
\textbf{g,} Target HPV$^{-}$ probability across editing steps (\textbf{left}) and source and counterfactual class probabilities (\textbf{right}). Target probability increases from 0.002 to 0.955; the dotted line marks the 0.5 decision threshold.
}

\label{fig:progressive}
\end{figure}

Counterfactual explanations of whole-slide predictions require morphological interventions that can extend beyond the bounded field of view of a local image generator. SCALE achieves this through progressive overlapping-window editing, which coordinates successive local generations through shared spatial context (\figref{fig:progressive}a,e). This strategy allows target-associated morphology to be propagated across increasingly large tissue regions while maintaining continuity between neighboring edited and unedited areas.

Progressive editing links these local operations across a shared spatial grid. {Each local generation window operates on a $256\times256$-pixel tile, and the $2048\times2048$-pixel regions used for systematic evaluation were therefore partitioned into an $8\times8$ conditioning grid of such tiles ($8\times256=2048$). This grid was traversed using overlapping local generation windows (\figref{fig:progressive}d,e).} 
Each {newly} generated {tile} 
was incorporated into the evolving tissue image and used as spatial context for subsequent interventions, extending the morphological changes across the selected region {(details on progressive editing in \secref{sec:progressive})}.

We compared progressive editing with an independent-window baseline using matched target concepts, editing locations and generation windows. {In the baseline, each window was generated independently and the outputs were assembled by hard stitching. \figref{fig:progressive}b shows that the independent-window baseline produced visible seams and abrupt discontinuities in tissue architecture at tile boundaries, whereas progressive editing yielded seamless, spatially coherent transitions in which generated tiles blended naturally with their neighbors and with the surrounding unedited tissue}.

Averaged across nine task directions, progressive editing produced lower mean absolute RGB change and lower excess border discontinuity {(defined in \secref{sec:boundary})} than the independent-window baseline (\figref{fig:progressive}c). {At a selected-tile budget of 48, mean absolute RGB change was 49.39 for progressive editing versus 63.72 for the independent-window baseline, a reduction of approximately 22\%.} {Excess border discontinuity was 3.70 for progressive editing versus 18.92 for the baseline, corresponding to an approximately 80\% reduction in boundary discontinuity.} These results indicate that our SCALE's progressive editing reduces overall pixel alteration and boundary artifacts relative to independent-window editing. 

In the representative HPV$^{+}$-to-HPV$^{-}$-targeted intervention, attention-guided selection and progressive window traversal extended editing across the source region (\figref{fig:progressive}d--f). The region-level HPV$^{-}$ probability increased from 0.002 to 0.955, crossing the decision threshold during editing (\figref{fig:progressive}g). {This result shows that SCALE can successfully flip a region-level prediction by coordinating local edits into a much larger, visually coherent, counterfactual region, laying the foundation for whole-slide counterfactual generation (as shown in \secref{sec:counter-whole}).} 

\section{Discussion}\label{sec:discussion}

We developed SCALE, a pathology-grounded framework for explaining whole-slide model predictions through spatially controlled morphological counterfactuals. Across nine counterfactual directions spanning six cancer types and multiple pathological endpoints, SCALE generated recognizable target-associated morphological changes that consistently shifted model predictions toward the intended targets (e.g., from HPV$^{+}$ to HPV$^{-}$), thereby counterfactually explain the original model prediction (e.g., HPV$^{+}$). Counterfactual responses depended on both the local strength and spatial extent of the intervention, while attention from the model being explained identified tissue regions in which morphological changes were most relevant to the prediction. SCALE's progressive editing further propagated localized interventions across large tissue regions and toward whole-slide scale while limiting unnecessary image perturbation and boundary artifacts. Together, these results demonstrate that with SCALE's counterfactual explanations, whole-slide model behavior can be examined through explicit changes in tissue morphology rather than localization alone.

A central contribution of SCALE is to move pathology model interpretation from association toward controlled morphological intervention. {Attention-based approaches can identify tissue regions associated with a prediction \citep{lu2021dataefficient}, while representation-based methods can identify interpretable morphological concepts and other cancer-related features \citep{le2024learning,qin2026clearhpv,su2025interpretable}.} These approaches reveal where predictive information is concentrated or which morphological patterns are associated with a model output, but do not directly test how changing those patterns alters the prediction. {Slide-level counterfactual approaches \citep{kaczmarzyk2024hippo} construct counterfactual slides by removing, adding or replicating existing patch embeddings, without synthesizing any pathological images; they are not generative methods and therefore cannot generate counterfactual explanations. 
{On the other hand, alongside broader advances in diffusion guidance \citep{wang2026inference, yang2026crossdomain},} generative counterfactual methods can produce prediction-altering histopathology images \citep{zigutyte2026counterfactual,munoz2026histoxplaining} but remain task-specific and limited to single-tile generation.
{Direct manipulation of individual interpretable latent features has also been explored \citep{kim2026picasso};} in our setting, such isolated SAE-feature steering frequently produced off-target morphology or image corruption. 
SCALE closes this gap by, for the first time, unifying WSI-level counterfactual reasoning with interpretable, spatially controlled morphological generation, enabling direct tests of how specific morphological changes alter whole-slide predictions.

More importantly, morphological counterfactuals provide a direct way to assess whether model decisions are grounded in pathologically meaningful evidence. When the model's prediction on the counterfactual image shifts in the intended direction in response to histologically plausible changes in recognizable tissue structures, the counterfactual provides evidence about the morphology the model relies on for its decision. 
{Conversely, a substantial prediction shift without coherent changes in diagnostically relevant morphology may indicate reliance on spurious visual cues, such as staining intensity or background texture, whereas an unchanged prediction despite histologically appropriate alterations may reveal discrepancies between model-learned features and established diagnostic criteria.}
{Such investigations could help pathologists identify model failure modes and inform subsequent model refinement.}
{Beyond evaluating model trustworthiness, recurring counterfactual morphological patterns could help identify candidate morphological markers associated with tumor status, histological grade or molecular status, providing a basis for further clinical and biological validation.}
Blinded assessment by six pathologists across 30 cases spanning HPV status, PRAD growth patterns and COAD provided complementary support for these morphological interventions. SCALE achieved the highest plausibility win rate among the evaluated methods, and readers selected its counterfactuals over source images as more consistent with the requested target morphology in nearly all paired assessments. {These findings support the morphological plausibility and clinical interpretability of SCALE's counterfactuals, providing a strong foundation for pathologists to critically investigate AI predictions and explore disease-associated morphology.}

Counterfactual explanation therefore complements conventional predictive performance by providing an additional basis for evaluating how a model reaches its decisions and, ultimately, for assessing model trustworthiness. Blinded assessment by six pathologists across 30 HPV-status, PRAD growth-pattern, and COAD provided complementary support for these morphological interventions. SCALE achieved the highest plausibility win rate among the evaluated methods, and readers selected its counterfactuals over source images as more consistent with the requested target morphology in nearly all paired assessments. These findings support both the relative morphological plausibility of SCALE and its ability to introduce target-associated changes (i.e., counterfactual explanations) recognizable to pathologists.

SCALE uses a reusable morphological concept space to define these counterfactual interventions across diverse pathology tasks. Rather than learning a separate representation for each prediction problem, as also explored in medical image generation \citep{chen2026foundation}, the same visually grounded concepts can be queried to identify target-associated morphology and reused as intervention targets across cancer types and endpoints. This makes the counterfactual images interpretable in morphological terms rather than as an arbitrary perturbation in representation space. The intervention experiments further revealed distinct aspects of model sensitivity: model responses depended both on how strongly target-associated morphology was expressed locally and on how broadly it was distributed across tissue. Attention provided a complementary spatial dimension by identifying where these concept-guided interventions were most relevant to the model prediction. Together, these capabilities allow counterfactuals to probe not only which morphology matters, but also how its expression and spatial distribution influence model behavior.

Spatial scalability is particularly important in whole-slide pathology because prediction-relevant morphology may extend across tissue regions far larger than the field of view of a local image generator. Our SCALE's Progressive editing allowed localized morphological interventions to be propagated across neighboring regions while maintaining continuity with previously generated and surrounding tissue. Compared with independent-window generation, this strategy substantially reduced boundary discontinuities and unnecessary image alteration, preventing seams and mismatched morphology from accumulating as the counterfactual expanded. The same procedure successfully scales from systematically evaluated \(2048\times2048\)-pixel regions to complete WSIs. Progressive editing therefore provides the spatial mechanism needed to preserve plausible morphology as counterfactual explanations extend from local regions toward whole-slide scale.

Several limitations remain. First, SCALE enables counterfactual generation across whole-slide spatial extents, but the current counterfactual intervention remains defined at the level of tissue tiles rather than individual histological structures. This provides scalable regional control but does not yet allow direct manipulation of specific glands, cells or subcellular features; future work could incorporate finer-grained or hierarchical representations to enable more precise morphological interventions while retaining whole-slide scalability. Second, the learned concept space depends on the underlying pathology foundation model, so poorly represented morphological attributes may be difficult to isolate or manipulate. Future work could explore alternative foundation models to obtain complementary or finer-grained concept spaces.

SCALE extends whole-slide interpretation from asking \emph{where} predictive evidence resides to testing \emph{which pathologically plausible tissue changes} would lead the same model to a different conclusion. By making these prediction-altering morphological changes explicit across spatial scales, the framework provides a direct means to examine whether model decisions are grounded in morphologically meaningful evidence; this is an important step toward the transparent and trustworthy deployment of digital pathology models in clinical practice \citep{kelly2019key}.


\section{Methods}\label{meth}

\subsection{Dataset preprocessing and feature extraction}

We used hematoxylin-and-eosin-stained whole-slide images (WSIs) from The Cancer Genome Atlas (TCGA) \citep{weinstein2013cancer} spanning multiple cancer types. The pan-cancer dataset was partitioned at the patient level into an SAE training set and a held-out test set. The SAE training set comprised 10,484 diagnostic WSIs from 8,641 patients across 32 TCGA cohorts and 25 anatomical sites, corresponding to 141,085,065 unique tissue tiles. The held-out test set comprised an additional 1,109 WSIs from 905 patients and 15,179,191 tissue tiles. In total, the dataset contained 11,593 WSIs from 9,546 patients and 156,264,256 unique tissue tiles.

Slides were processed at 20$\times$ magnification and partitioned into non-overlapping $256\times256$-pixel tiles. Tiles with insufficient tissue content, determined from the proportion of non-background pixels, were excluded before feature extraction. No stain or color normalization was applied to the image tiles, and no additional normalization was applied to the extracted foundation-model features before SAE training.\footnote{We also evaluated color normalization and observed no meaningful change in results, likely due to the foundation model's robustness to staining and color variation acquired from diverse pretraining data.} All slides from the same patient were assigned to the same partition to prevent patient-level information leakage.

Let $t_{m,i}$ denote tissue tile $i$ from slide $m$, where $i=1,\ldots,N_m$, with $N_m$ as the number of tiles for slide $m$. Each retained tile was encoded using the frozen \mbox{UNI2-h} pathology foundation model $f_{\mathrm{UNI2}}$ \citep{chen2024uni} to obtain a $D=1,536$-dimensional feature representation,
\begin{align}
x_{m,i}=f_{\mathrm{UNI2}}(t_{m,i}), \qquad x_{m,i}\in\mathbb{R}^{D}.
\end{align}
For slide $m$, the resulting tile features were stacked as
\[
X_m=[x_{m,1},\ldots,x_{m,N_m}]^\top\in\mathbb{R}^{N_m\times D}.
\]
The UNI2-h encoder remained frozen throughout SAE training, downstream model training, and counterfactual evaluation. SAE parameters and concept prototypes were learned exclusively from the SAE training partition; the held-out partition was \emph{not} used for SAE training, concept-prototype construction, or hyperparameter tuning. Task-specific downstream models were similarly developed using patient-disjoint training and test partitions.

\subsection{End-to-end counterfactual generation}

The end-to-end SCALE workflow follows the seven stages summarized in \figpanelref{fig:framework}{c}. Steps 1--3 define \emph{what} morphology should be changed for a specified alternative prediction. In step 1, a sparse autoencoder (SAE) decomposes pathology foundation-model representations into sparse latent components capturing recurrent morphological patterns. In step 2, these components are then grounded in image space using their top-activating tissue tiles, forming a bank of morphological concepts reusable across cancer types (more details in~\secref{sec:sae}). In step 3, for a specified source-to-target counterfactual direction, concept-class associations are used to identify the concept most strongly associated with the target label (e.g., HPV$^+$); this concept, together with its latent prototype, serves as the morphological target for counterfactual intervention (more details in~\secref{sec:concept_class}).

Once the target morphology has been identified, step 4 translates this concept into a feature-space intervention. The SAE representation of each selected tile is steered toward the latent prototype of the target concept and then decoded back into the pathology foundation-model feature space. This produces a modified feature representation that preserves the original tile representation while shifting it toward the selected morphological alternative, e.g., shifting from HPV$^+$ to HPV$^-$ (more details in~\secref{sec:steering}). 

Step 5 performs this feature-space intervention in image space. Specifically, the steered feature from step 4 is provided as conditioning (input) to a frozen feature-conditioned diffusion model, which generates the corresponding morphological counterfactual explanation image. The locations selected for intervention are prioritized using attention from the same MIL model whose prediction is being explained (\figpanelref{fig:framework}{b}), thereby determining \emph{where} the target-associated morphology is introduced. Surrounding tissue is retained as spatial context during generation so that the induced morphological change remains localized to the intended region (more details in~\secref{sec:steering}).

When the requested counterfactual extends beyond the field of view of a single image generation window, step 6 propagates these local interventions across larger tissue regions, potentially to the entire slide. SCALE applies overlapping generation windows progressively across the selected locations, carrying the continuously updated tissue image forward so that previously generated morphology provides context for subsequent edits. This progressive procedure determines \emph{how} localized morphological changes are extended across larger spatial extents and toward whole-slide scale while maintaining continuity between neighboring regions (more details in~\secref{sec:progressive}). 

Finally, step 7 closes the counterfactual explanation loop. The generated counterfactual tissue is re-encoded using the same frozen tile encoder and returned to the same MIL model used for the original prediction. The resulting change in target-class probability measures how the model responds to the introduced morphological alternative (i.e., the counterfactual explanation). Thus, SCALE connects the morphology selected for intervention, the tissue location at which it is introduced, and its spatial realization in image space to the resulting change in model prediction.

\subsection{Sparse autoencoder and concept prototypes}\label{sec:sae}

To obtain a sparse representation of pathology features, we trained a sparse autoencoder (SAE) \citep{cunningham2023sparse} on tile-level UNI2-h representations from the SAE training partition. For notational simplicity, we write \(x_i\) for the feature of tile \(i\), suppressing its slide index $m$. Given an input feature \(x_i\in\mathbb{R}^{D}\), the SAE encoder produced an overcomplete latent representation
\begin{align}
z_i
=
\operatorname{ReLU}\!\left(
W_{\mathrm{enc}}x_i+b_{\mathrm{enc}}
\right),
\qquad
z_i\in\mathbb{R}^{L},
\end{align}
where \(L=12,288>D=1,536\). The ReLU activation constrained SAE activations to be non-negative. The corresponding decoder reconstructed the pathology feature as
\begin{align}
\hat{x}_i
=
W_{\mathrm{dec}}z_i+b_{\mathrm{dec}}.
\end{align}

The SAE was optimized using a reconstruction objective with an \(\ell_1\) sparsity penalty,
\begin{align}
\mathcal{L}_{\mathrm{SAE}}
=
\frac{1}{B}
\sum_{i=1}^{B}
\left\|x_i-\hat{x}_i\right\|_2^2
+
\lambda_{\mathrm{SAE}}
\frac{1}{B}
\sum_{i=1}^{B}
\left\|z_i\right\|_1,
\end{align}
where \(B\) denotes the number of tile features in a training batch and \(\lambda_{\mathrm{SAE}}\) controls the strength of sparsity regularization. Because \(z_i\) is non-negative, the \(\ell_1\) term penalizes total activation magnitude and encourages each tile to be represented by a relatively small subset of active SAE dimensions.

Each SAE latent dimension was treated as a candidate morphological concept. For concept \(j\), training tiles were ranked according to their activation \(z_{i,j}\) in the corresponding latent dimension. We used the \(K_{\mathrm{proto}}=10\) highest-activating tiles to represent each concept, denoting the tile set as 

\begin{align}
\mathcal{T}_j
=
\operatorname{TopK}
\left(
\{z_{i,j}\}_{i=1}^{N_{\mathrm{tr}}};\,
K_{\mathrm{proto}}
\right),
\end{align}
where $N_{\mathrm{tr}}$ is the number of tiles in the SAE training partition and $\operatorname{TopK}(\cdot\,;K)$ returns the indices of the $K$ largest values in the set. The corresponding image tiles were retained as visual examples of the morphology associated with concept $j$.

For counterfactual steering, we additionally constructed a latent prototype for each concept $j$ by averaging the complete SAE codes of these top-activating tiles,
\begin{align}
p_j
=
\frac{1}{|\mathcal{T}_j|}
\sum_{i\in\mathcal{T}_j}
z_i,
\qquad
p_j\in\mathbb{R}^{L}.
\end{align}
Thus, the scalar activation \(z_{i,j}\) was used to identify tiles strongly expressing concept \(j\), whereas \(p_j\) retained the joint latent configuration characteristic of those representative tiles. We used this complete latent prototype $p_j$ as the steering target because the morphology represented by a concept may depend on the configuration of multiple SAE dimensions rather than on a single latent activation alone. This choice was further evaluated against direct single-neuron steering in an ablation experiment. Concept prototypes were constructed exclusively from the SAE training partition and remained fixed during downstream counterfactual experiments.

\subsection{Attention-based MIL and concept-class association}\label{sec:concept_class}

Task-specific attention-based multiple-instance learning (MIL) models were trained for breast, colorectal, kidney, lung, head-and-neck and prostate pathology using patient-disjoint training and test partitions. The evaluated tasks comprised BRCA, COAD, KIRC and LUAD normal-versus-tumor classification; HNSCC HPV-status classification; and a three-class PRAD growth-pattern task comprising GP1--3, GP4 and GP5. For PRAD, GP1--3 denotes an experimental grouping of discrete, well-formed glands based on the historical Gleason patterns 1--3; patterns 1 and 2 are no longer routinely assigned in contemporary clinical grading \citep{epstein2016contemporary}.

For slide $m$ with $N_m$ tiles, the set of tile features $\{x_{m,i}\}_{i=1}^{N_m}$, $x_{m,i}\in\mathbb{R}^{d}$,
was treated as a bag of instances. The corresponding MIL model assigned normalized attention weights \(a_{m,i}\) to individual tiles, where
\[
a_{m,i}\geq0,
\qquad
\sum_{i=1}^{N_m}a_{m,i}=1,
\]
and aggregated the tile representations into a slide-level representation by attention-weighted averaging,
\[
h_m=\sum_{i=1}^{N_m}a_{m,i}\,x_{m,i},
\qquad
h_m\in\mathbb{R}^{d},
\]
{which was passed to a classification head to obtain the slide-level prediction. Attention weights therefore provided a model-derived measure of the relative contribution of each tissue tile to the slide-level prediction.}

To identify morphological concepts associated with each downstream class, SAE activations were combined with attention weights from the corresponding trained MIL model. Let \(z_{m,i,j}\) denote the activation of concept \(j\) for tile \(i\) from slide \(m\). We defined the attention-weighted slide-level activation of concept \(j\) as
\begin{align}
q_{m,j}
=
\sum_{i=1}^{N_m}
a_{m,i}z_{m,i,j}.
\label{eq:q}
\end{align}
Because $\sum_{i=1}^{N_m}a_{m,i}=1$, $q_{m,j}$ is a weighted mean over tiles and its scale does not depend on the number of tiles $N_m$, so slides with larger size do not receive larger scores.

For class \(c\), let \(\mathcal{M}_{c}^{\mathrm{train}}\) denote the set of training slides with label \(c\). The mean attention-weighted activation of concept \(j\) within class \(c\) was
\begin{align}
\mu_j^{(c)}
=
\frac{1}{|\mathcal{M}_{c}^{\mathrm{train}}|}
\sum_{m\in\mathcal{M}_{c}^{\mathrm{train}}}
q_{m,j}.
\label{eq:mu}
\end{align}


{Averaging within each class in Eq.~\eqref{eq:mu} normalizes by the class size $|\mathcal{M}_{c}^{\mathrm{train}}|$, so classes with more training slides do not disproportionately influence concept-association estimates. Together, Eqs.~\eqref{eq:q} and \eqref{eq:mu} therefore reduce bias from unequal slide size and unequal class size.}

For a counterfactual intervention from source class \(c_{\mathrm{src}}\) to target class \(c_{\mathrm{tgt}}\), we defined the concept-class association contrast as
\begin{align}
\Gamma_j^{(c_{\mathrm{tgt}}\leftarrow c_{\mathrm{src}})}
=
\mu_j^{(c_{\mathrm{tgt}})}
-
\mu_j^{(c_{\mathrm{src}})}.
\end{align}
Positive values therefore indicate concepts with stronger attention-weighted expression in the target class than in the source class. The target-associated concept used for steering was then selected as
\begin{align}
j^*
=
\operatorname*{arg\,max}_{j}
\Gamma_j^{(c_{\mathrm{tgt}}\leftarrow c_{\mathrm{src}})}.
\end{align}

Note that all concept-class associations above were computed using training slides only, and the selected concept \(j^*\) for each source-to-target intervention direction was fixed before evaluation on held-out slides. Thus, the pan-cancer SAE concept bank remained unchanged across a wide range of downstream tasks, while the morphological concept selected for counterfactual intervention was determined separately for each task and intervention direction.

\subsection{Attention-guided concept steering}\label{sec:steering}

For counterfactual intervention, non-background tissue tiles within the evaluated region were ranked according to attention weights from the corresponding MIL model. Let \(\mathcal{V}_m\) denote the set of non-background tissue tiles in slide \(m\), with \(N_m^{\mathrm{tissue}} = |\mathcal{V}_m|\), and let \((i_1,\ldots,i_{N_m^{\mathrm{tissue}}})\) denote these tiles ordered such that

\begin{align}
a_{m,i_1}\geq a_{m,i_2}\geq\cdots\geq a_{m,i_{N_m^{\mathrm{tissue}}}}.
\end{align}

For an editing budget of $k$ tiles, the attention-guided edit set was

\begin{align}
\mathcal{I}_{\mathrm{edit},m}^{(k)}
=
\{i_1,\ldots,i_k\}.
\end{align}

For the random-selection baseline, the same number of tiles was sampled uniformly at random from $\mathcal{V}_m$. The target concept, steering parameters, diffusion generator and editing budget were otherwise held fixed, isolating the effect of spatial selection.

Let $j^*$ denote the selected target-associated concept and $p_{j^*}$ its latent prototype. For notational simplicity, we omit the slide index $m$ below. For each selected tile $i\in\mathcal{I}_{\mathrm{edit}}$, the original SAE code $z_i$ was interpolated toward the target prototype,

\begin{align}
z_i'
=
z_i+s\left(p_{j^*}-z_i\right)
=
(1-s)z_i+sp_{j^*},
\qquad
0\leq s\leq1,
\end{align}

where $s$ denotes the SAE steering strength. Setting $s=0$ leaves the original latent representation unchanged, whereas $s=1$ replaces it with the target prototype.

The steered latent representation was decoded back into pathology-feature space and interpolated with the original UNI2-h feature,

\begin{align}
x_i'
=
(1-\beta)x_i
+
\beta
\left(
W_{\mathrm{dec}}z_i'
+
b_{\mathrm{dec}}
\right),
\qquad
0\leq\beta\leq1,
\end{align}

where $\beta$ controls the contribution of the decoded steered feature to the final conditioning representation. Unless otherwise specified, counterfactual experiments used steering strength
\(s=0.9\) and feature-mixing coefficient \(\beta=0.1\). Steering strength was
varied only in the intervention-strength analysis. Tiles outside the edit set retained their original features,

\begin{align}
x_i'=x_i,\qquad
i\notin\mathcal{I}_{\mathrm{edit}}.
\end{align}

The resulting features were returned to their original spatial locations to form the edited conditioning grid $F_{\mathrm{edit}}$, with each grid cell corresponding to one spatially aligned $256\times256$-pixel pathology tile. In other words, each returned UNI2-h feature vector $x_i'$ will be fed into a pathology diffusion model~\citep{yellapragada2025pixcell} to generate one $256\times256$-pixel pathology tile, laying the foundation for~\secref{sec:progressive} below.

\subsection{Progressive diffusion-based counterfactual generation}\label{sec:progressive}

Counterfactual images were generated using a frozen feature-conditioned pathology diffusion model \citep{yellapragada2025pixcell} that operates on a spatial \(4\times4\) grid of conditioning features, with each cell corresponding to a \(256\times256\) tile. To extend generation beyond this native field of view, we developed a progressive overlapping-window procedure that coordinates successive local interventions through shared spatial context.

{Let $I^{(0)}$ denote the original source region. Each conditioning-grid cell corresponded to one spatially aligned $256\times256$-pixel pathology tile. Generation proceeded through overlapping $4\times4$-tile local windows ($1024\times1024$ pixels), each containing a fully edited $2\times2$ center surrounded by a blended peripheral region (\figref{fig:progressive}a). By traversing these overlapping windows across the tissue, editing was extended over the full $8\times8$ global conditioning grid that constitutes each $2048\times2048$-pixel region used for large-region evaluation (\figref{fig:progressive}e).}

{For each active window, SAE-steered conditioning features $x_i'$ were applied only to selected tiles (\figref{fig:progressive}d) when they entered the $2\times2$ editable center (\figref{fig:progressive}a). Unselected cells within the editable center and all cells in the peripheral region retained their original features $x_i$. } 
Concept-guided intervention was therefore confined to the intended spatial locations, while surrounding conditioning features provided local morphological context.

At progressive step \(r\), the diffusion generator operated on the corresponding region of the current composite image \(I^{(r-1)}\) (e.g., a large \(2048\times2048\)-pixel region or the whole slide). Generated pixels within the central editable region were incorporated fully into the composite image. Within the peripheral region, generated output was blended with the existing composite using 10\% newly generated content and 90\% existing content, while pixels outside the active window remained unchanged. The resulting image \(I^{(r)}\) was then carried forward to the next overlapping window.

Because each subsequent window operated on the progressively updated image rather than the original source region, previously generated tissue provided spatial context for later interventions. Windows were traversed until all selected tiles had entered a central editable region, allowing localized counterfactual changes to be propagated across larger tissue extents while maintaining continuity with neighboring tissue.

To evaluate extension beyond the standardized large \(2048\times2048\)-pixel regions, we applied the same progressive procedure to complete COAD WSIs. Attention-prioritized tissue locations were edited through successive overlapping local windows on a continuously updated slide image. Each generation step retained the same bounded \(4\times4\) conditioning window used for large-region experiments. Counterfactual morphology was then progressively propagated across spatially distributed tissue regions while preserving the surrounding slide context. 

For the independent-window baseline, we used the same diffusion generator, conditioning features, window geometry, traversal sequence and spatial composition rule. The only difference was that each local window was generated from the corresponding region of the original source image \(I^{(0)}\), rather than from the progressively updated composite \(I^{(r-1)}\). Previously generated tissue therefore did not provide context for subsequent windows.

\subsection{Counterfactual evaluation}

We evaluated SCALE at two spatial scales: whole-slide counterfactual editing as the primary evaluation and fixed-region counterfactual editing as a complementary analysis. Both settings used held-out samples from the designated source class of each intervention direction and measured how progressively increasing the spatial extent of counterfactual intervention affected predictions from the corresponding task-specific multiple-instance learning (MIL) model.

\textbf{Whole-slide counterfactual evaluation.}
The primary evaluation included 134 distinct whole-slide images (WSIs) from 101 TCGA cases across nine counterfactual task directions. For each WSI, all non-background tissue tiles were considered eligible for intervention. Attention weights from the same task-specific MIL model whose prediction was being explained were used to rank eligible tissue tiles in descending order, prioritizing prediction-relevant regions for counterfactual editing.

To evaluate the effect of spatial intervention extent across WSIs containing different numbers of tissue tiles, editing budgets were expressed as percentages of the total eligible tissue tiles rather than fixed tile counts. Let $\mathcal{V}_m$ denote the set of eligible tissue tiles in slide $m$, with $|\mathcal{V}_m|$ representing the total number of eligible tiles. For an editing extent of $p\%$, the corresponding editing budget was defined as
\begin{align}
k_m(p) = \left\lceil \frac{p}{100}|\mathcal{V}_m| \right\rceil,
\end{align}
with $k_m(0)=0$. The cumulative edit set $\mathcal{I}_{\mathrm{edit},m}^{(p)}$ contained the $k_m(p)$ highest-ranked eligible tissue tiles. Edit sets were nested, such that tiles selected at smaller editing percentages remained included at larger intervention extents.

Counterfactual images were generated at the selected spatial locations using the progressive overlapping-window editing procedure, allowing concept-guided morphological interventions to extend across the WSI while preserving surrounding tissue context. At each editing extent, the generated counterfactual tissue tiles were re-encoded using the frozen UNI2-h encoder, and their features replaced the corresponding original tile features in the whole-slide feature bag. The resulting feature bag was then evaluated using the original task-specific MIL model to measure the target-class probability after intervention.

\textbf{Region-level counterfactual evaluation.}
As a complementary evaluation, we examined counterfactual editing within fixed $2048\times2048$-pixel tissue regions. Candidate regions from multiple held-out source-class slides were first evaluated using the corresponding task-specific MIL model. Only regions for which the pre-intervention prediction matched the source slide label were retained. From this eligible pool, 30 regions were randomly sampled per slide for counterfactual evaluation.

Within each selected region, only non-background tissue tiles were eligible for intervention. Attention weights from the same MIL model were used to rank these tiles and determine their editing order.

To evaluate increasing intervention extent within the fixed regions, tiles were cumulatively edited using
\begin{align}
k \in \{1,8,16,32,48\},
\end{align}
where $k$ denotes the absolute number of edited tiles. The edit sets were nested, with tiles selected at smaller editing budgets retained at larger budgets. Let $\mathcal{V}_{m,r}$ denote the eligible tissue tiles in region $r$ of slide $m$, and let
$\mathcal{I}_{\mathrm{edit},m,r}^{(k)}
\subseteq\mathcal{V}_{m,r}$
denote the cumulative set of the $k$ highest-ranked tissue tiles. The corresponding generated counterfactual region is denoted by $I_{m,r,k}^{\mathrm{cf}}$.

\textbf{Whole-slide prediction assessment.}
For both evaluation settings, generated counterfactual tiles were re-encoded using the same frozen UNI2-h encoder used for the original WSI. Let
\begin{align}
X_m=[x_{m,1};\dots;x_{m,N_m}]
\in\mathbb{R}^{N_m\times d}
\end{align}
denote the original whole-slide feature bag of slide $m$, containing $N_m$ tile features as introduced in Section~\ref{sec:concept_class}.

For the whole-slide evaluation, let $X_m^{\mathrm{cf},(p)}$ denote the counterfactual feature bag obtained after editing $p\%$ of the eligible tissue tiles. For the region-level evaluation, let $X_{m,r}^{\mathrm{cf},(k)}$ denote the whole-slide feature bag after replacing features corresponding to the $k$ edited tiles within region $r$. In both settings, features outside the respective edit sets remained unchanged.

The original task-specific MIL model was then applied to the resulting counterfactual feature bag. For target class $y_{\mathrm{tgt}}$, the whole-slide target-class probability at editing extent $p$ was defined as
\begin{align}
\pi_{\mathrm{tgt},m}(p)
=
P\left(
y_{\mathrm{tgt}}
\mid X_m^{\mathrm{cf},(p)}
\right).
\end{align}

Analogously, region-level interventions were evaluated by applying the same MIL model to $X_{m,r}^{\mathrm{cf},(k)}$. This procedure quantified the extent to which counterfactual morphological interventions shifted the predictions of the original model being explained, without retraining or modifying its parameters.

The target-directed probability change relative to the original WSI was defined as
\begin{align}
\Delta\pi_{\mathrm{tgt},m}(p)
=
\pi_{\mathrm{tgt},m}(p)
-
\pi_{\mathrm{tgt},m}(0).
\end{align}
Positive values indicate that the counterfactual intervention increased the model's predicted probability of the requested target class. The same evaluation principle was applied to region-level interventions using the corresponding counterfactual whole-slide feature bags.

The benchmark comprised nine source-to-target intervention directions: BRCA normal-to-tumor, COAD normal-to-tumor, KIRC normal-to-tumor, LUAD normal-to-tumor, HNSCC HPV-positive-to-HPV-negative, HNSCC HPV-negative-to-HPV-positive, PRAD GP1--3-to-GP4, PRAD GP4-to-GP5 and PRAD GP1--3-to-GP5. For each direction, the corresponding MIL model, target-associated concept and steering parameters were fixed before evaluation on held-out data.

To separately examine the effects of local steering strength and spatial editing extent, we performed two additional controlled analyses on representative tissue regions. For the steering-strength analysis, the spatial intervention region was held fixed while the SAE steering strength was varied from 0\% to 100\%. For the editing-extent analysis, steering parameters were held fixed while the fraction of non-background tissue tiles selected for intervention was increased from 0\% to 100\%. Target-class probability and mean activation of the selected target concept were recorded after each intervention.

\subsection{Visual perturbation and boundary analysis} \label{sec:boundary}

Visual perturbation was measured between spatially aligned source and counterfactual regions. Let $\Omega$ denote the evaluated image domain. The mean absolute channel-wise RGB difference was defined as

\begin{align}
\mathcal{E}_{\mathrm{RGB}}(k)
=
\frac{1}{3|\Omega|}
\sum_{u\in\Omega}
\left\|
I_k^{\mathrm{cf}}(u)-I^{(0)}(u)
\right\|_1,
\end{align}
{where $I^{(0)}(u)\in\mathbb{R}^3$ denotes the RGB intensity vector of the source region at pixel $u$, and $I_k^{\mathrm{cf}}(u)\in\mathbb{R}^3$ denotes the RGB intensity vector of the counterfactual region at the same pixel location after editing $k$ tiles. The factor $1/3$ averages the absolute difference over the three color channels.} 
RGB intensities were represented on the $[0,255]$ scale, so $\mathcal{E}_{\mathrm{RGB}}$ is reported in RGB intensity units on the same scale. Lower values indicate less pixel-level deviation from the source image.

To quantify discontinuities introduced where neighboring generation windows met, let $\mathcal{B}$ denote the set of adjacent pixel pairs $(u,v)$ lying on opposite sides of the evaluated window boundaries. For image $J$, boundary discontinuity was defined as

\begin{align}
\operatorname{BD}(J)
=
\frac{1}{3|\mathcal{B}|}
\sum_{(u,v)\in\mathcal{B}}
\left\|
J(u)-J(v)
\right\|_1.
\end{align}
{where $J(u)\in\mathbb{R}^3$ and $J(v)\in\mathbb{R}^3$ denote the RGB intensity vectors of $J$ at the two pixels of a boundary pair, so that $\operatorname{BD}(J)$ is the mean absolute channel-wise intensity difference across window boundaries.} Excess boundary discontinuity relative to the spatially aligned source image was

\begin{align}
\mathcal{E}_{\mathrm{border}}(k)
=
\operatorname{BD}\!\left(I_k^{\mathrm{cf}}\right)
-
\operatorname{BD}\!\left(I^{(0)}\right).
\end{align}

The same boundary locations were used when evaluating progressive and naive independent-window generation. A value of zero indicates boundary discontinuity equal to that already present in the source image, whereas positive values indicate additional discontinuity introduced by generation. Lower $\mathcal{E}_{\mathrm{RGB}}$ and $\mathcal{E}_{\mathrm{border}}$ therefore indicate less visual perturbation and lower excess boundary discontinuity, respectively.

For each editing budget, visual metrics were calculated separately for each task direction and then averaged across the nine intervention directions. Shaded uncertainty bands denote the standard error of the mean across task directions ($n=9$), with task direction serving as the unit of cross-task summary.

\bibliography{custom}
\bibliographystyle{plainnat}


\end{document}